\documentclass{ifacconf}

\usepackage{graphicx}      
\usepackage{natbib}        
\usepackage{amsmath}
\usepackage{bm}
\usepackage{amsfonts}
\usepackage{array}

\makeatletter
\def\endfigure{\end@float}
\def\endtable{\end@float}
\makeatother

\usepackage[caption=false]{subfig}
\usepackage{etoolbox}
\usepackage{threeparttable}

\AtBeginEnvironment{figure}{\setlength{\captionwidth}{0.8\linewidth}}
\AtBeginEnvironment{table}{\setlength{\captionwidth}{0.8\linewidth}}
\begin{document}
\begin{frontmatter}

\title{A Dynamic Penalty Function Approach for Constraint-Handling in Reinforcement Learning} 

\author[First]{Haeun Yoo} 
\author[Second]{Victor M. Zavala}
\author[First]{Jay H. Lee}

\address[First]{Department of Biomolecular and Chemical Engineering, Korea Advanced Institute of Science and Technology, 291 Daehak-ro, Yuseong-gu, Daejeon, 34141, Republic of Korea (e-mail: haeungd, jayhlee@ kaist.ac.kr).}
\address[Second]{Department of Chemical and Biological Engineering, University of Wisconsin-Madison, Madison, WI 53706, USA (e-mail: victor.zavala@wisc.edu)}

\begin{abstract}                
Reinforcement learning (RL) is attracting attention as an effective way to solve sequential optimization problems that involve high dimensional state/action space and stochastic uncertainties. Many such problems involve constraints expressed by inequality constraints. This study focuses on using RL to solve constrained optimal control problems. Most RL application studies have dealt with inequality constraints by adding soft penalty terms for violating the constraints to the reward function. However, while training neural networks to learn the value (or Q) function, one can run into computational issues caused by the sharp change in the function value at the constraint boundary due to the large penalty imposed. This difficulty during training can lead to convergence problems and ultimately lead to poor closed-loop performance. To address this issue, this study proposes a dynamic penalty (DP) approach where the penalty factor is gradually and systematically increased during training as the iteration episodes proceed. We first examine the ability of a neural network to represent a value function when uniform, linear, or DP functions are added to prevent constraint violation. The agent trained by a Deep Q Network (DQN) algorithm with the DP function approach was compared with agents with other constant penalty functions in a simple vehicle control problem. Results show that the proposed approach can improve the neural network approximation accuracy and provide  faster convergence when close to a solution.
\end{abstract}


\end{frontmatter}

\section{Introduction}
Sequential decision-making problems such as scheduling, planning, and control involve constraints that should be respected. When the state space system model is available, the optimization problem can be formulated as follows:
\begin{subequations}
\begin{align}\label{Eq:Objective}
\min_{[\bm{u_k}]_{k=0}^T} \; & \sum\limits_{k=0}^T l(\bm{x}(k), \bm{u}(k)) \\
\textrm{s.t.}\; & \bm{x}(k+1) = f(\bm{x}(k), \bm{u}(k)), \label{Eq:state_transition_model}\\
& g_i(\bm{x}(k), \bm{u}(k)) \leq 0, \; i = 1, ... \upsilon \label{Eq:inequality_const}\\
& \bm{x}(0) = \bm{x}_0,\label{Eq:initial_state}\\
& \underline{\bm{x}} \leq \bm{x}(k) \leq \overline{\bm{x}}, \label{Eq:state_bound}\\
& \underline{\bm{u}} \leq \bm{u}(k) \leq \overline{\bm{u}}, \label{Eq:input_bound}
\end{align}
\end{subequations}
where $k$ is the time index, $\bm{x} \in \mathbb{R}^n$ is the state vector, $\bm{u} \in \mathbb{R}^m$ is the control input, and $l(\bm{x,u})$ is a cost function. By solving the above problem at each time instant for the given state (estimate) as the initial state $\bm{x}_0$ and implementing the resulting $\bm{u}_0$ as the control action, one can effectively implement an optimal feedback policy. This is a basic idea behind the popular model predictive control (MPC) method. However, the model inevitably has some mismatch with the real system due to model errors, disturbances, noise, etc. The basic (deterministic) MPC formalism does not allow one to consider various uncertainties in the optimization, but instead compensate for their effect passively through the feedback. This can lead to a performance loss including constraint violations. To address this problem, reinforcement learning (RL) has been suggested as an alternative to MPC in the process system engineering field (see \cite{yoo2020reinforcement, petsagkourakis2020reinforcement, shin2019reinforcement}).

An RL agent finds the optimal policy by exploring different parts of the state space through simulations or by probing the real system to learn the approximate value of each state. Using the Bellman optimality principle, the value estimate can be improved iteratively, with the aim of eventually identifying the optimal value (\cite{bellman1966dynamic}). In RL, the control problem is formulated as a Markov decision process, which is composed of the state space ($\bm{S}$) and state ($\bm{x} \in \bm{S}$), action space ($\bm{A}$) and action ($\bm{u} \in \bm{A}$), the state transition model, and the reward function ($r: S \times A \to R$) (\cite{puterman2014markov}). The state transition model  \eqref{Eq:state_transition_model} is used in this study but this can be generalized to specify probabilistic transitions and/or the effect of stochastic noise terms.

The agent gets a certain reward (or penalty) for each state visited and the summation of the discounted rewards (with a discount factor of $\gamma \in [0,1]$) from current state to the terminal state ($\sum_{k=t}^T\gamma^{k-t}r(x(k))$) is the value of the state. The value function, $V(x)$, is the function that maps the state to the expectation of the return value, which implies the long-term \textit{"value"} of the state, and Q-function $Q(x,a)$, is the expected return value of an state and action pair. In the MDP formulation, the action limits of \eqref{Eq:input_bound} can always be satisfied by considering only the feasible actions. Meanwhile, the state bounds of \eqref{Eq:state_bound} and state/input constraints \eqref{Eq:inequality_const} cannot generally be satisfied in a strict sense as they may become infeasible. To address this problem, policy gradient RL methods for constrained MDP have been proposed based on a trust region policy optimization algorithm (\cite{achiam2017constrainedpolicyopt, petsagkourakis2020constrained} and a Lyapunov-based approach (\cite{chow2019lyapunov}). However, these methods are only applicable within the policy gradient algorithms. Hence, we instead focus on the penalty approach, where violations are penalized with terms added to the reward function along with the original objective function \eqref{Eq:Objective}. This approach is more general as it can be applied to not only policy-based methods, but also other value-based RL methods.

The form of penalty function can be a uniform constant, as shown in \cite{yang2020unipen, zhang2020unipen}, a linear (1-norm) function of the magnitude of the violation as shown in \cite{ma2019linearpen}, or a logistic function as shown in \cite{pan2020logpen, modares2016logpen}. The penalty is typically chosen to be large enough in order to prevent constraint violation. On the other hand, an infinite penalty  (as used in the barrier method for solving constrained optimization) can lead to infeasibilities and thus will not be used here. When we use a high uniform penalty value or a steep linear penalty function, the training of neural network (NN), a popular choice for representing the value (or Q) function in RL, can give convergence problems as will be demonstrated later with a simple example. On the other hand, a small penalty term can lead to unnecessary constraint violations, i.e., violations even when the constraint can be satisfied. A penalty function that strikes a right balance is needed, but this is not easy to decide {\em a priori}. 

In this study, we propose an approach that updates the penalty factor as the training proceeds, in order to achieve a stable training result with less approximation error and to get a so-called "sufficiently feasible" policy (\cite{deb2000efficient}). We will refer to this approach as the DP approach. A similar approach has been studied to update parameters during iterations in evolutionary optimization strategies (\cite{kramer2010evolutionreview, joines1994dynamicpenalty}). Also, \cite{lin2012penup} suggested to gradually increase the penalty value in RL-based control to address the numerical difficulty approximating a steep function, but they do not provide any detailed procedure or systematically analyze the effect of varying the penalty parameter.

The paper is structured as follows: Section 2 describes the DP function design with constraints aggregation using the Kreisselmeier-Steinhauser (KS) function and penalty factor update rules. In Section 3, we use a simple 1-dimensional function approximation example to examine the regression accuracy achieved with several penalty functions including the DP function. In Section 4, the proposed DP function is tested on a simple vehicle control example and the results are summarized. Section 5 concludes the paper.

\section{Dynamic Penalty (DP) Function} \label{sec:DynamicPenalty}

In this work we use a constraint aggregation method to design an unbiased penalty for constraint violations. A systematic rule for updating the penalty factor is described for an application to any RL algorithm. 

\subsection{Constraint Aggregation}

Let us assume that we have $\upsilon$ inequality constraints including the inequality constraints on the state space boundary. The RL agent gets a penalty value for each constraint violation. A constraint aggregation is needed for unbiased constraint handling when multiple constraints exist. The KS function is a widely used constraint aggregation method for gradient-based optimization (\cite{kreisselmeier1980KS}). Although RL is not a gradient-based optimization algorithm, we can effectively use this method with a high aggregation parameter, $\rho$, making the error from the aggregation as small as possible (\cite{poon2007adaptiveKS}). For constraints $g_i(x_t, u_t) \leq 0, i = 1, ..., \nu$, the aggregate function $KS[\bm{g(x, u)}]$ is:
\begin{equation} \label{eq:KS}
\begin{split}
KS&[\bm{g(x,u)}] = \\
&g_{max}(\bm{x,u})+\frac{1}{\rho} \ln [\sum^\nu_{i=1}e^{\rho(g_i(\bm{x,u})-g_{max}(\bm{x,u}))}].
\end{split}
\end{equation}
where $g_{max}$ is the maximum of all constraints evaluated at the given state-action pair.

\subsection{DP Function}

\begin{figure}
\begin{center}
\includegraphics[trim=120 150 120 150, clip, width=7.0cm]{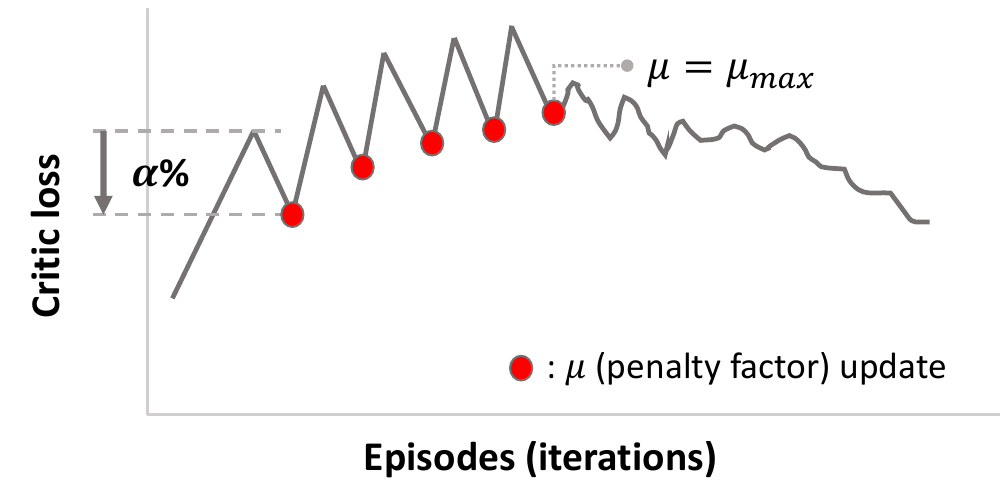}    
\caption{Illustration of penalty update rule} 
\label{fig:PenaltyUpdate}
\end{center}
\end{figure}

Using the KS function, the reward function with DP term can be defined as follows:
\begin{subequations}
\begin{align}
    R(\bm{x,u})&=l(\bm{x,u})+p(\bm{x,u}|\mu) \\
    p(x,u|\mu)&=\mu \cdot KS[\bm{g(x,u)}]\bm{1}_{KS[\bm{g(x,u)]>0}} \label{3b}
\end{align}
\end{subequations}
where $\mu$ is the penalty factor which defines the slope of the penalty function. The right-hand side of \ref{3b} is calculated only when the aggregate constraint is violated. The DP function approach starts with a low value of $\mu$, which facilitates the initial training of the NN, and gradually increases it to a large value by multiplying the DP update parameter $c$ so that the slope becomes large to prevent constraint violations (Fig. \ref{fig:PenaltyUpdate} illustrates this approach). The procedure can be described as:

\begin{enumerate}
  \item Set $\mu := \mu_{min}$.
  \item When the loss value of the NN is less than $(100-\alpha) \%$ of the maximum loss value after the parameter update, set $\mu := c \mu$ where $c$ is some constant larger than 1 \label{step2} \label{step2}
  \item Repeat step (\ref{step2}) until $\mu \geq \mu_{max}$
  \item After reaching $\mu \geq \mu_{max}$ set $\mu := \mu_{max}$
  \item Continue the training until an optimal policy is found.
\end{enumerate}

\section{Value Function Regression Test} \label{sec:fitting}

\subsection{1-Dimensional Example}
A feed-forward NN is the most widely used function approximator for the value function or Q-function approximation. The agent uses the output value or the gradient of the NN to update the optimal policy. Therefore, accuracy of the approximation is important for a fast and stable agent training. We tested the regression accuracy with a priori fixed uniform penalty and linear penalty, which are the popular choices for the penalty function against the proposed dynamically varied penalty function. The test value function $V(x)$ is chosen as follows:

\begin{equation}
    V(x) = 1 + \cos{(0.5x)}+0.05(x-1)(x+2) + p(x).  \label{eqn:1dobjective}
\end{equation}

We assumed that this problem has inequality constraints as $-5 \leq x \leq 5$ and the penalty functions are assigned as in table \ref{tb:1dpenalty} where the $\mu_{min}=0.1$, $\mu_{max}=50$ and $c=2$. This artificially chosen value function is not exactly in the same form as in the optimal control problem where the value should be calculated with respect to the cumulative rewards. Nevertheless, this analysis should be useful to understand the tendencies of the NN approximation during the agent training with the different penalty terms. 

\begin{table}[hb]
\begin{center}
\caption{Parameters for the penalty function} \label{tb:1dpenalty}
\begin{tabular}{l|c}
\hline
\textbf{Penalty function} & $\bm{p(x)}$\\
\hline
Uniform penalty & $50 \cdot \bm{1}_{x<-5,x>5}$ \\
\hline
Linear penalty & $50((-5-x)\bm{1}_{x<-5}+(x-5)\bm{1}_{x>5})$\\
\hline
DP & $\mu((-5-x)\bm{1}_{x<-5}+(x-5)\bm{1}_{x>5})$\\
\hline
\end{tabular}
\end{center}
\end{table}

To examine the NN training performance within an RL environment, the training environment is set similarly as in RL. In every episode, 20 $x$ points are randomly sampled from -10 to 10 and the environment calculates the $R(x)$ value ($R(x)=l(x)+p(x)$) with the assigned objective function \eqref{eqn:1dobjective} and the penalty function (table \ref{tb:1dpenalty}). Then, 20 samples are stored in a replay buffer and the NN is trained with a batch of 64 data randomly sampled from the replay buffer up to 500 episodes. 

\subsection{Results} \label{subsec:1dResult}

\begin{figure*}[htb]
    \centering
    \subfloat[][Uniform penalty]{
    \includegraphics[width=0.3\textwidth]{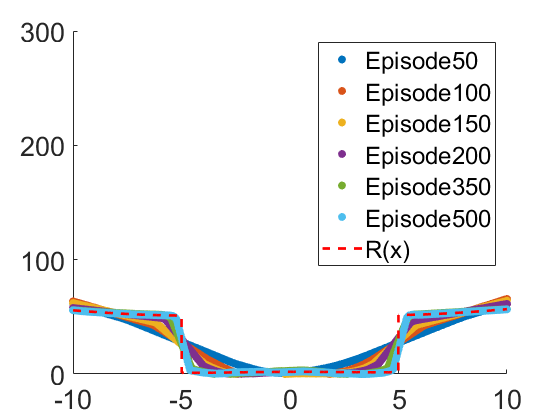}
    \label{fig:1duni}
    }
    \subfloat[][Linear penalty]{
    \includegraphics[width=0.3\textwidth]{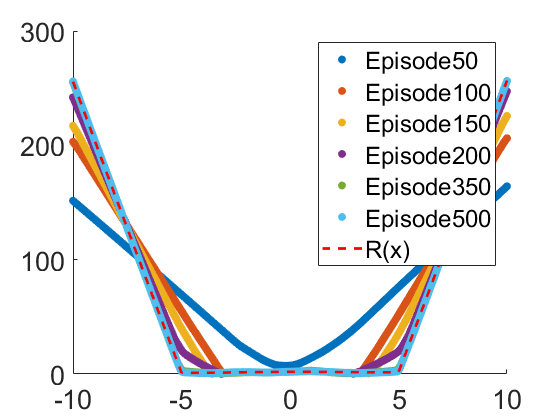}
    \label{fig:1dlin}
    }
    \subfloat[][DP]{
    \includegraphics[width=0.3\textwidth]{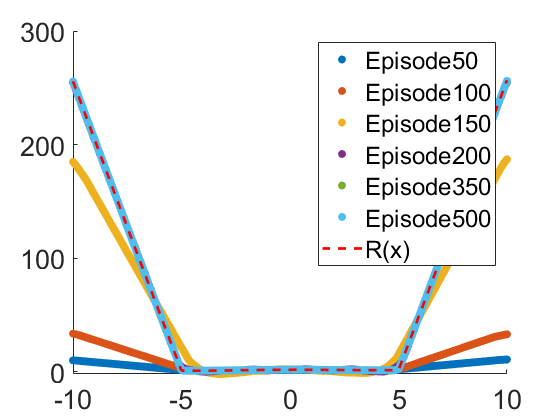}
    \label{fig:1dup}
    }
    \caption{NN approximation results according to episodes progression}
    \label{fig:1d}
\end{figure*}

\begin{figure}
    \centering
    \includegraphics[width=0.35\textwidth]{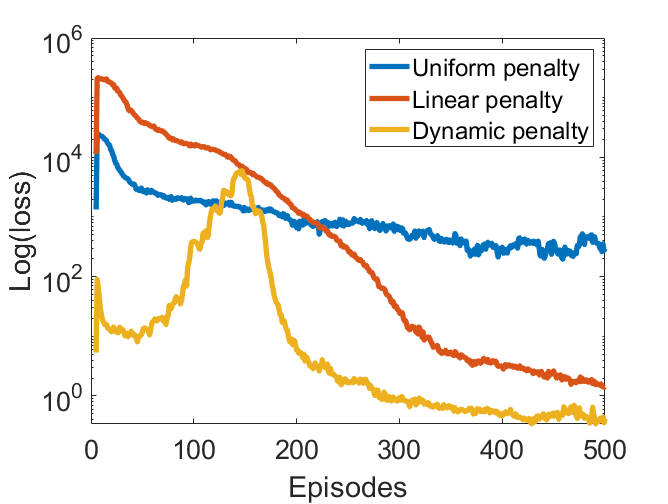}
    \caption{NN loss according to episodes progression}
    \label{fig:1dloss}
\end{figure}

Fig. \ref{fig:1d} shows the output value of the NN during the training process. As shown in Fig. \ref{fig:1duni}, the NN approximation performance is poor at the boundary points, which causes a bias in the estimate inside the feasible region. The linear penalty case also gives a biased estimate inside the feasible region during the training and the maximum loss function is much higher than that of the DP case, as shown in Fig. \ref{fig:1dloss}. With the proposed scheme of dynamically varying the penalty during traning, the function is well approximated near the constraint boundary, which can reduce the bias and loss. Fig. \ref{fig:1dloss} shows that the loss of NN with the DP function approach is increased during the penalty factor update period but is decreased drastically afterward. The final loss is the lowest in the case of DP, which implies the best approximation performance. From this simple test, we see the potential benefit of using a dynamically varied penalty in training the NN to learn the value function with a large penalty function.

\section{Case Study} \label{sec:casestudy}

\subsection{Vehicle Control Example} \label{subsec:simple}

To further examine the effectiveness of the DP approach, a simple vehicle control problem described in \eqref{eq:simple} is solved with RL. We modified the problem from \cite{grune2017simplevehicle} to reduce the size of the feasible region and make the starting point uncertain as described in \eqref{eq:simple_initial}. The objective is to minimize the running cost \eqref{eq:simple_obj}. $x_1(k)$ is the position of a vehicle at time k, $x_2(k)$ is its velocity and $u(k)$ is its acceleration. 
\begin{subequations}
\label{eq:simple}
\begin{equation}
    \min \; l(x,u)=x_1^2+u^2 \label{eq:simple_obj}
\end{equation}
\begin{equation}
    \textrm{s.t. }\; \left( \begin{array}{c} \dot{x}_1(k)\\ \dot{x}_2(k) \end{array} \right) = \left( \begin{array}{c} x_2(k) \\ u(k) \end{array} \right)
\end{equation}
\begin{equation}
    -1 \leq x_1(k) \leq 1 \label{eq:simple_x1cons}
\end{equation}
\begin{equation}
    -0.25 \leq x_2(k) \leq 1 \label{eq:simple_x2cons}
\end{equation}
\begin{equation}
    -0.25 \leq u(k) \leq 0.25
\end{equation}
\begin{equation}
    x_0 \sim U((-1, 0.8), (-0.8, 1)) \label{eq:simple_initial}
\end{equation}
\end{subequations}
We applied the Deep-Q-Network (DQN) algorithm which uses a deep NN to approximate the Q-function and chooses the $\epsilon$-greedy policy (\cite{mnih2013DQN}). For the training, we defined the discrete action space $u \in \{-0.25, -0.2375, -0.225, ..., 0.2375, 0.25\}$ and each episode was set for a time duration of 20min with 1min control interval. The deep NN consisted of three hidden layers with 64 nodes of the ReLU activation function. For constraint aggregation, \eqref{eq:simple_x1cons} and \eqref{eq:simple_x2cons} are decomposed as shown in \eqref{eq:simple_g} and aggregated to a $KS[\bm{g}]$ function of \eqref{eq:KS} with $\rho=50$.
\begin{equation}
\label{eq:simple_g}
    \left( \begin{array}{l} 
    g_1: -1-x_1 \leq 0 \\
    g_2: x_1-x \leq 0 \\
    g_3: -0.25 -x_2 \leq 0 \\
    g_4: x_2 -1 \leq 0 \end{array} \right)
\end{equation}

The DP was updated when the loss became lower than 30\% of the maximum loss ($\alpha = 70$) and the penalty factor was doubled in each update ($c=2$), starting from $\mu_{min}=0.05$ up to $\mu_{max}=20$. The value for the uniform penalty function and the penalty factor for the linear penalty were both set to 20 as shown in table \ref{tb:simplepenalty}. The training comprised 2000 episodes and the initial point was randomly selected inside the feasible region to start each episode. To test the agent’s behavior during training, we tested the policy after every 100 episodes without exploration. In addition, for a rigorous comparison, we trained a set of 100 agents in the environment set with 100 different random seeds, which introduced randomness to exploration, NN initialization, buffer sampling, etc.

\begin{table}[hb]
\begin{center}
\caption{Penalty function types} \label{tb:simplepenalty}
\begin{tabular}{l|c}
\hline
\textbf{Penalty function} & $\bm{p(x)}$\\
\hline
Uniform penalty & $20 \cdot \bm{1}_{KS(\bm{g})>0}$ \\
\hline
Linear penalty & $20 \cdot KS(\bm{g}) \cdot \bm{1}_{KS(\bm{g})>0}$\\
\hline
DP & $\mu \cdot KS(\bm{g}) \cdot \bm{1}_{KS(\bm{g})>0}$\\
\hline
\end{tabular}
\end{center}
\end{table}

\subsection{Results} \label{subsec:simresult}

Table \ref{tb:casestudy1} shows that the agents trained with the uniform penalty and the linear penalty could find sufficiently feasible policies only in 42 and 48 cases out of the 100 random cases, respectively. Here, ‘sufficient feasible policy’ means that the policy does not violate all the constraints and achieves a sufficient low cost value. The agent trained with the uniform penalty incurred the highest average cost. Also notable is that the agent could find feasible policies in 82 cases, but almost a half of them were unable to achieve sufficiently low cost. According to table \ref{tb:casestudy2}, the uniform penalty seems favorable for a rapid training, but the final performance is not satisfactory. These results are likely due to the interference of the bias in the estimate inside the feasible region. The agent trained with the linear penalty function performed poorly due to having inaccurate approximate values for the value function during training. This consequently resulted in the policy being updated in a wrong direction, or converging to sub-optimal policies.

\begin{figure}
    \centering
    \includegraphics[width=0.35\textwidth]{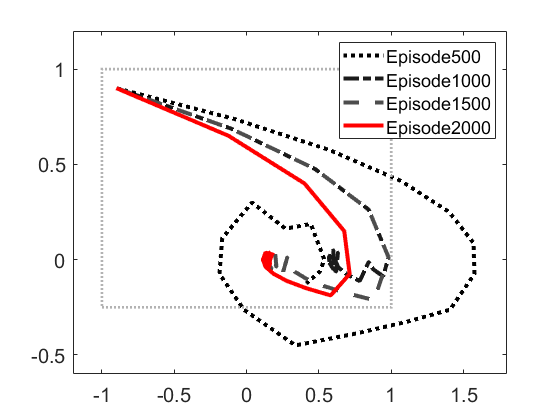}
    \caption{Trajectories in the training process of the agent trained with a DP function. The dotted rectangle represents the feasible region.}
    \label{fig:dynamictpenaltytraj}
\end{figure}

The use of the DP approach was effective in finding a sufficient feasible policy rapidly and consistently. The agent trained with the DP approach could find sufficient feasible policies in 83 cases, and the average cost of that cases was lower than the results of the agent trained with other penalty functions. We also calculated the average degree of violation by averaging the positive KS function values from the converged infeasible policy. As shown in Table \ref{tb:casestudy1}, when the agent was trained with the DP approach, the degree of violation of the infeasible policies was much lower compared to the other approaches. This means that even the infeasible policies found in the 7 cases, nearly satisfied the constraints. Table \ref{tb:casestudy2} also shows that sufficiently feasible policies were found within 1500 episodes in 39 cases, which is comparable to the training with the uniform penalty. Fig. \ref{fig:dynamictpenaltytraj} shows the policy improvement during training with the DP function.

\begin{table*}[htbp]
\centering
\captionsetup{width=.8\linewidth}
\caption{Summary of agent training results}
\begin{tabular}{l|c|c|c}
\hline
\textbf{Penalty function} & \textbf{Uniform penalty} & \textbf{Linear penalty} & \textbf{DP}\\
\hline
\textbf{Finding feasible policy/100} & 82 & 62 & 93 \\
\hline
\textbf{Finding sufficient feasible policy*/100} & 42 & 48 & 83\\
\hline
\textbf{Average cost of the sufficient feasible policy} & 3.56 & 3.49 & 3.36 \\
\hline
\textbf{Average degree of violation} & 0.948 & 2.051 & 0.137 \\
\hline
\end{tabular}
\begin{tablenotes}
\centering
\item[a] $*$'sufficient feasible policy' means the policy does not violate all the constraints and achieves a sufficient low cost value
\end{tablenotes}
\label{tb:casestudy1}
\end{table*}

\begin{table}[htbp]
\caption{Finding sufficient feasible policy until each episodes}
\begin{center}
\begin{tabular}{l|c|c|c|c}
\hline
\textbf{Penalty function} & \textbf{500} & \textbf{1000} & \textbf{1500} & \textbf{2000}\\
\hline
Uniform penalty & 0&14&15&13\\
\hline
Linear penalty & 2&8&16&22\\
\hline
DP & 2&8&29&44\\
\hline
\end{tabular}
\label{tb:casestudy2}
\end{center}
\end{table}

\section{Conclusions}

To solve constrained optimal control problems with RL, inequality constraints imposed on the state should be translated into a penalty term in the reward function. However, when the agent is trained with a priori set large uniform or linear penalty function, inaccurate NN approximations can result leading to convergence problems and poor eventual performance. To address this, we propose an approach that varies the penalty function during training in order that NNs be trained stably and rapidly. Our results show that the DP approach can reduce the approximation loss not only during the training but also at the end of the training. The DP approach was also found effective in the simple vehicle control problem tested. We trained the agent with different penalty forms, and the DP scheme showed the best performance in finding sufficient feasible policies with lowest average cost. In addition, the average degree of violation was lowest when the agent was trained with the DP function. The proposed approach can be applied to any RL algorithm and can help the agent that uses NN as a function approximator to be trained efficiently to represent a function that includes a steep penalty term for constraint violation. For future work, we will further analyze the effectiveness of the DP function approach for cases with model and exogenous uncertainties. 

\begin{ack}
This research was supported by Korea Institute for Advancement of Technology (KIAT) grant funded by the Korea Government(MOTIE)(P0008475, The Competency Development Program for Industry Specialist).  Victor Zavala acknowledges support from the members of the Texas-Wisconsin-California control consortium. 
\end{ack}

\bibliography{ifacconf}             

\end{document}